%% file: kg_llm_bench.tex
\documentclass{article} 
\usepackage[preprint]{colm2025_conference}

\usepackage{microtype}
\usepackage{hyperref}
\usepackage{url}
\usepackage{booktabs}
\usepackage{float}

\usepackage{lineno}

\definecolor{darkblue}{rgb}{0, 0, 0.5}
\hypersetup{colorlinks=true, citecolor=darkblue, linkcolor=darkblue, urlcolor=darkblue}

\usepackage{times}
\usepackage{latexsym}

\usepackage[T1]{fontenc}

\usepackage[utf8]{inputenc}

\usepackage{microtype}

\usepackage{inconsolata}

\usepackage[symbol, bottom]{footmisc}
\usepackage{authblk}
\usepackage{graphicx}
\usepackage{subcaption}
\usepackage{amsmath}
\usepackage{amssymb}
\usepackage[ruled,vlined]{algorithm2e}
\usepackage{booktabs}
\usepackage{multirow}
\usepackage{makecell}
\usepackage{longtable}
\usepackage{rotating}
\usepackage{enumitem}
\usepackage[most]{tcolorbox}
\tcbuselibrary{listings}
\title{\texttt{KG-LLM-Bench:} A Scalable Benchmark for Evaluating LLM Reasoning on Textualized Knowledge Graphs}

\newcommand{\coauthmark}{\textsuperscript{*}}

\author[1]{Elan Markowitz\coauthmark\textsuperscript{$\dagger$}}
\author[2]{Krupa Galiya\coauthmark\textsuperscript{$\ddagger$}}
\author[1,3]{Greg Ver Steeg}
\author[1]{Aram Galstyan}
\affil[1]{University of Southern California}
\affil[2]{Independent Researcher}
\affil[3]{University of California, Riverside}

\begin{document}

\ifcolmsubmission
\linenumbers
\fi

\maketitle

\footnotetext[1]{Denotes equal contribution}
\footnotetext[2]{esmarkow@usc.edu}
\footnotetext[3]{krupagaliya@gmail.com}

\begin{abstract}

Knowledge graphs have emerged as a popular method for injecting up-to-date, factual knowledge into large language models (LLMs). This is typically achieved by converting the knowledge graph into text that the LLM can process in context. While multiple methods of encoding knowledge graphs have been proposed, the impact of this textualization process on LLM performance remains under-explored. We introduce KG-LLM-Bench, a comprehensive and extensible benchmark spanning five knowledge graph understanding tasks, and evaluate how different encoding strategies affect performance across various base models. Our extensive experiments with seven language models and five textualization strategies provide insights for optimizing LLM performance on KG reasoning tasks.

\end{abstract}

\section{Introduction}

\begin{figure*}[b]
    \centering
    \includegraphics[width=\linewidth]{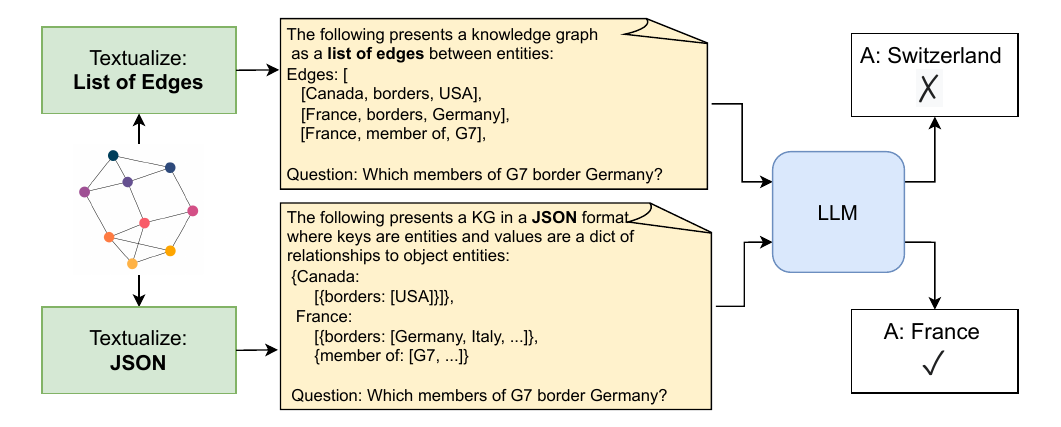}
    \caption{Different formats for graph textualization can result in highly varied performance on downstream tasks. }
    \label{fig:example}
\end{figure*}

The integration of knowledge graphs (KGs) with large language models (LLMs) has emerged as an important approach for enhancing contextual understanding in AI systems \citep{kau2024combiningknowledgegraphslarge}. Knowledge graphs are large structured databases of factual information that encode real world entities and their relationships. Recent surveys have highlighted the complementary nature of LLMs (static, unstructured, opaque, general) and KGs (dynamic, structured, interpretable, specific) \citep{Pan2023LargeLM, Zhu2023LLMsFK, Yang2023ChatGPTIN, Li2024surveygraphmeetslarge, Pan_2024, Fan2024graphmachinelearningera}. This synergy has driven research on specialized architectures and algorithms for their integration \cite{Luo2023ReasoningOG, Wen2023MindMapKG, Markowitz2024TreeofTraversalsAZ}. Most of these approaches rely on converting the KG to a readable text format suitable for LLM processing.

However, many of these algorithms give little consideration to the specific method of KG textualization \citep{fatemi2023talklikegraphencoding, chen2024textspacegraphfoundationmodels, yu2024g2tllmgraphtotreetextencoding}. The most common approach simply encodes the KG as a list of edges in the form (source entity label, relation, object entity label). It is assumed that any approach will be equally effective and that using the same format ensures fair model comparison \citep{guo2023gpt4graphlargelanguagemodels,wang2024languagemodelssolvegraph}.

In this work, we challenge these assumptions and demonstrate that textualization strategy significantly impacts performance. Our experiments show that choosing the right strategy can improve overall benchmark performance by up to 17.5\% absolute difference, with even larger gains on specific tasks. Figure \ref{fig:example} illustrates an example of the problem we are trying to analyze.

Our contributions are:

\begin{enumerate}[]
    \item A scalable and extensible benchmark for analyzing how LLMs process and understand in-context knowledge graphs with five tasks covering important KG reasoning capabilities. 
    \item Experiments covering five different textualization strategies using seven different popular LLM models, resulting in new insights and best practices. 
    \item Experiments with pseuodnyms showing that LLMs do not heavily rely on memorized information when processing in-context knowledge graphs (overall difference of 0.2\%). 
    \item A public release of the benchmark and framework so that it can be rapidly expanded. 
\end{enumerate}

\section{Background}

\textbf{Knowledge Graphs} are large structured databases that store factual associations as edges in a graph. They come in many varieties from general knowledge \citep{Vrandei2014WikidataAF, Lehmann2015DBpediaA} to domain-specific variants such as Finance, Geology, and Medicine
\citep{Liu2019AnticipatingSM,Zhu2017IntelligentLF,Choi2019InferenceOB,Farazi2020KnowledgeGA}. 
We formally define a source knowledge graph $\mathcal{K}=(\mathcal{E}, \mathcal{R}, \mathcal{T})$ where $\mathcal{E}$ is the set of entities, $\mathcal{R}$ is the set of relation types, and $\mathcal{T}$ is the set of edges of the form $(s, r, o)\in\mathcal{E}\times \mathcal{R} \times \mathcal{E}$ e.g., (`Inception', `director', `Christopher Nolan').

We can define a subgraph of $\mathcal{K}$ as $G = (G_\mathcal{E}, G_\mathcal{R}, G_\mathcal{T})$ where $G_\mathcal{E}\subseteq\mathcal{E}$, $G_\mathcal{R}\subseteq\mathcal{R}$, and $G_\mathcal{T}\subseteq\mathcal{T}$.

\textbf{Large Language Models} can learn from information passed into their context window. This is used in retrieval augmented generation to produce more accurate LLM responses \cite{Lewis2020RetrievalAugmentedGF}. We can define the generation process using LLM $\pi$ that responds to a query $q$ using context text $c$:
\begin{equation}
\label{eq:llm}
    \hat{y} = \pi(c, q)
\end{equation}
where $\pi$ generates a response $\hat{y}$. This context can include any text-format data, including various text encodings for knowledge graphs (Fig \ref{fig:example}).

\section{Related Work}

\textbf{Benchmarks for Graphs Reasoning} \ \ Recent work has extensively evaluated LLM understanding of graph-structured data. Many benchmarks focus on simple graphs rather than knowledge graphs \citep{guo2023gpt4graphlargelanguagemodels, tang2024grapharenabenchmarkinglargelanguage, yuan2024gracorebenchmarkinggraphcomprehension}. Other research has expanded to text-space graph foundation models \citep{chen2024textspacegraphfoundationmodels} and hypergraphs \citep{feng2024graphslargelanguagemodels}. Particularly relevant to our work, \citet{fatemi2023talklikegraphencoding} evaluates how different natural language presentations affect graph understanding, and was later extended to learned graph encoders \citep{Perozzi2024LetYG}.

\textbf{KG Question Answering (KGQA) Benchmarks}\ \ The KGQA field has produced several key benchmarks, including QALD-10 \citep{usbeckqald}, 2WikiMultiHop \citep{xanh2020_2wikimultihop}, and MetaQA \citep{zhang2017variationalMetaQA}. While HotpotQA \citep{Yang2018HotpotQAAD} is not KG-grounded, it is still often used for knowledge grounded evaluation. Recent work like CLR-Fact \citep{zheng2024clrfactevaluatingcomplexlogical} evaluates LLMs on complex logical query answering \citep{Arakelyan2020ComplexQA, Galkin2024AFM}.

More focused studies have examined specific KG processing capabilities in LLMs, including KG completion \citep{Yao2023ExploringLL}, construction \citep{zhu2024llmsknowledgegraphconstruction}, causal reasoning \citep{kim2024causalreasoninglargelanguage}, and trustworthiness enhancement \citep{sui2024knowledgegraphsmakelarge}. While some work has explored KG formatting, such as RDF Turtle parsing \citep{Frey2023BenchmarkingTA} and natural language presentations \citep{dai2024largelanguagemodelsunderstand}, our work presents the first comprehensive evaluation of textualization strategies.

\textbf{KG-Grounded Models and Algorithms}\ \ Recent approaches to grounding LLMs with knowledge graphs include Think-on-Graph \citep{Sun2023ThinkonGraphDA} and MindMap \citep{Wen2023MindMapKG}. Tree-of-Traversals \citep{Markowitz2024TreeofTraversalsAZ} enables test-time search over KG reasoning paths. Alternative approaches like graph-to-tree text encoding \citep{yu2024g2tllmgraphtotreetextencoding} focus on specialized encoding strategies.

\textbf{LLM Reasoning and Long Context Models}\ \ Two emerging research directions could significantly impact KG reasoning capabilities. Test-time reasoning models like OpenAI's o1/o3 and DeepSeek's R1 \citep{DeepSeekAI2025DeepSeekR1IR} enable using enhanced computational resources at inference-time. Meanwhile, advances in long context models \citep{Chen2023ExtendingCW, Lieber2024JambaAH, Hooper2024KVQuantT1, Munkhdalai2024LeaveNC} allow processing of larger knowledge graphs. Our benchmark is positioned to evaluate both these developments in the context of graph reasoning.

\section{KG-LLM-Bench Framework}

\begin{figure*}
    \centering
    \includegraphics[width=\linewidth]{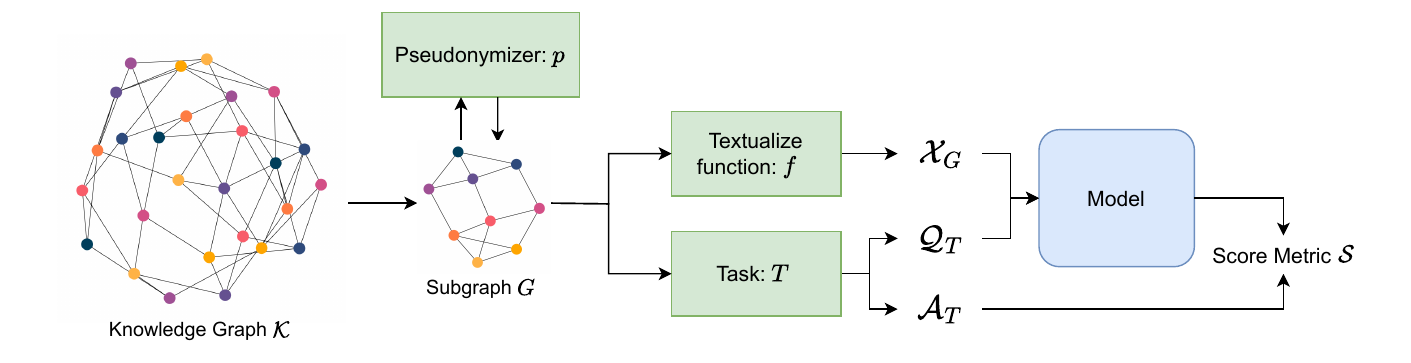}
    \caption{Framework for KG-LLM-Bench.}
    \label{fig:framework}
\end{figure*}

This section details the methodology of KG-LLM-Bench, which evaluates LLMs on knowledge graph question answering tasks. In summary, the LLM answers task-specific questions based on a KG subgraph $G$, with responses evaluated against predefined scoring criteria. Figure \ref{fig:framework} provides an overview of our framework.

\subsection{Text Representation of KG}  
We define a set of textualization functions $\mathcal{F}$ that convert the structured knowledge graph $G$ into a textual representation $x_G\in \mathcal{W}^*$, where $\mathcal{W}^*$ represents the set of all possible text strings:
\begin{equation}
\label{eq:textualize}
    x_G = f(G)
\end{equation}
where $f\in\mathcal{F}$. The specific textualization functions are detailed in Sec \ref{sec:TextualizationFunction}.

\subsection{Query Construction}
\label{sec:queryconstruct}
We design a set of tasks where each task $T$ (e.g., triple retrieval, shortest path) can be used to generate queries $q$ and answers $a$ for graph $G$:
\begin{align}
    q = \mathcal{Q}_T(G)\\ 
    a = \mathcal{A}_T(G, q)
\end{align}
where $\mathcal{Q}_T$ formulates the natural language query and $\mathcal{A}_T$ generates the corresponding answer. These are stochastic functions but are implemented with fixed seeds to ensure deterministic behavior.

\subsection{Model Generation and Evaluation}
The LLM $\pi$ generates an answer $\hat{y}=\pi(x_G,q)$ using the textualized graph as context. We evaluate this against the ground truth $a$ using a scoring function $\mathcal{S}$:
\begin{equation}
\label{eq:scoring}
    s = \mathcal{S}(\hat{y}, a) \in \{0,1\}
\end{equation}
where $s$ indicates correctness. While $\mathcal{S}$ is customizable, we use exact match in our experiments. 


\subsection{Optimizing for an LLM}

We can consider optimizing the textualization choice for a given model as an optimization of the expected performance of the LLM over the distribution of tasks, possible graph contexts, and questions and answers:

\begin{equation}
    \max_{f\in\mathcal{F}}\mathbb{E}_{T,G, \mathcal{Q}, \mathcal{A}}\Bigg[\mathcal{S}\Big(\pi\big(f(G), \mathcal{Q}_T(G), \mathcal{A}_T(G, \mathcal{Q}_T(G))\big)\Big)
    \Bigg]
\end{equation}

\subsection{Sampling graphs}
\label{sec:sampling}
We sample graphs $G \sim \textit{subgraph}(\mathcal{K})$ using seed entities, sampling radius, and max edges parameters. For each seed entity, we first sample ego-graphs containing all edges within the specified radius. An ego-graph for entity $e$ with radius $r$ is defined as $EgoGraph(e,r) = $
\begin{equation}
    \left\{ t=(s,r,o) | d(e,s) \leq r, d(e,o) \leq r, t\in \mathcal{T}\right\}
\end{equation}
where $d$ is the graph distance function. After combining ego-graphs, we apply a low-degree filter to remove single-edge entities, then randomly prune edges to meet the size constraint.

\subsection{Pseudonymization}
\label{sec:pseudonymization}
To ensure models rely solely on the provided knowledge graph rather than pre-trained knowledge, we introduce a pseudonymization function $p$ that maps entities to synthetic labels. Given a set of pseudonymized entity labels $\hat{\mathcal{E}}$, we create:
\begin{equation}
\label{eq:subgraph}
    \hat{G} = p(G, \hat{\mathcal{E}})
\end{equation}
where the pseudonymization creates the mapping $\left\{e: \hat{e} | e \in G_\mathcal{E}, \hat{e}\in\hat{\mathcal{E}}\right\}$ and applies it to $G$. 

For our experiments with historical country entities, we generate semantically appropriate pseudonyms using a combination of a name generator tool\footnote{https://www.name-generator.org.uk/} and LLM-generated names, filtering inappropriate or insulting samples.

\section{KG LLM Tasks}

We present five fundamental tasks in our benchmark, each chosen to evaluate distinct aspects of KG reasoning: retrieval, path-based reasoning, local aggregation, multi-hop aggregation, and global analysis. Together, these tasks provide a comprehensive evaluation of an LLM's ability to reason over knowledge graphs.

\subsection{Triple Retrieval Task}
The TripleRetrieval task tests an LLM's fundamental ability to verify the presence of relationships in graph $G$. This capability underlies all more complex graph reasoning tasks, as models must first accurately identify existing relationships before performing higher-order reasoning.

Questions are evenly split between positive and negative cases. Positive samples are drawn directly from edges $(s,r,o)\sim G_\mathcal{T}$. For negative samples, we create invalid edges by replacing either the source, relation, or object with alternatives ($s', o' \sim G_\mathcal{E}$ or $r'\sim G_\mathcal{R}$) such that the resulting edge does not exist in $G_\mathcal{T}$.

\subsection{Shortest Path Task}

The ShortestPath task evaluates a model's ability to find the shortest path between two entities in $G$, considering edges in either direction. This task is relevant as the shortest path between two entities is likely to represent the strongest association between those two entities. For instance, ``my brother's employer'' is a more direct and informative association than ``my mother's sister's nephew's employer''. Detailed implementation is provided in Appendix \ref{app:shortest_path}.

\subsection{Aggregation By Relation}

The AggByRelation task tests local aggregation from anchor nodes, a common requirement in real-world queries. For example, ``How many diplomatic relations does Uruguay have?'' requires aggregating connections from a specific entity.

Questions in this task take the form ``How many \{\textit{incoming/outgoing}\} relations of type \{\textit{relation type}\} does \{\textit{anchor entity}\} have?''. Since randomly sampling a relation type and direction and anchor entity most likely results in an aggregation over a single edge, we modify the approach to ensure variety in both the questions and answers. Details of this can be found in Appendix \ref{app:agg_by_relation}.

\subsection{Aggregation of Neighbor Properties}

The AggNeighborProperty task extends aggregation to two-hop paths, requiring more complex reasoning. Models must answer questions of the form ``How many of the directly connected entities to \{\textit{anchor entity}\} have an outgoing property of type \{\textit{relation}\} in the knowledge graph?''. Many real questions combine aggregation on multi-hop edges. For instance ``How many actors who starred in Inception have won Academy Awards?'' or ``How many universities that collaborate with Stanford University have research centers focused on artificial intelligence?''.

The task uses similar sampling approaches to AggByRelation (Appendix \ref{app:agg_neighbor_property}).

\subsection{Highest Degree Node by Direction}

The HighestDegree task tests global graph reasoning by identifying the entity with the most (incoming/outgoing/total) edges in $G$. The distinction between edge directions is significant. Since many textualization functions group edges with the same source, counting outgoing degree is a more local problem than counting incoming degree.

While more difficult global tasks could be proposed (e.g. graph isomorphism or connectivity statistics), we note that at the scale of graph we use, this task already proves to be relatively difficult.

\section{Experiments}

The following section describes the setup for our experiments. Our benchmark consists of five tasks (100 instances each, plus pseudonymized versions) and five textualization strategies, and we evaluate on seven LLMs.

\subsection{Data}

\input{tables/combined_dataset_table_and_best_format_table}

Our experiments use the Countries knowledge graph from WikiDataSets \citep{boschin_wikidatasets_2019} as our source graph $\mathcal{K}$. This knowledge graph is a subgraph related to historical countries derived from Wikidata \citep{Vrandei2014WikidataAF}.

The graph contains diverse relationship types covering geographical relations (e.g., borders), political relations (e.g., diplomatic relations), and temporal relations (e.g., followed by). In addition to the 49 core relations there are 162 attribute relations that connect countries to other types of entities such as languages or significant events. Table \ref{tab:dataset} summarizes the key statistics of the dataset.

When sampling subgraphs for our tasks, we follow the procedure outlined in Section \ref{sec:sampling}. The specific parameters were chosen to ensure reasonable questions could be generated for each task and that the subgraph is reasonable in terms of context size. Each question is asked over a subgraph with 200 edges. Full details are available in the appendix. We generate 100 sets of subgraph, question, and answer for each task.

\subsection{Textualization Strategies}
\label{sec:TextualizationFunction}
We evaluate five common textualization strategies for converting knowledge graphs into text:

\begin{enumerate}[noitemsep,nolistsep]
    \item \textbf{List-of-Edges}: A simple triple-based representation where each line contains a (subject, predicate, object) statement. Edges are presented in order by subject and relation.
    
    \item \textbf{Structured YAML}: A hierarchical representation using YAML syntax, grouping relationships by subject entities. 
    
    \item \textbf{Structured JSON}: Similar to YAML but using JSON syntax.
    
    \item \textbf{RDF Turtle}: A W3C standard format for representing RDF graphs, using prefixes and semicolons to group statements with the same subject. This format is commonly used in semantic web applications.
    
    \item \textbf{JSON-LD}: A JSON-based format for linked data that provides both human-readable structure and semantic web compatibility through the inclusion of contexts and URIs.
\end{enumerate}

Each format represents different tradeoffs between compactness, readability, and structure, allowing us to evaluate how these characteristics affect LLM performance. Details in Appendix \ref{app:text_formats}.

\begin{figure*}[t]
    \centering
    \includegraphics[width=\linewidth]{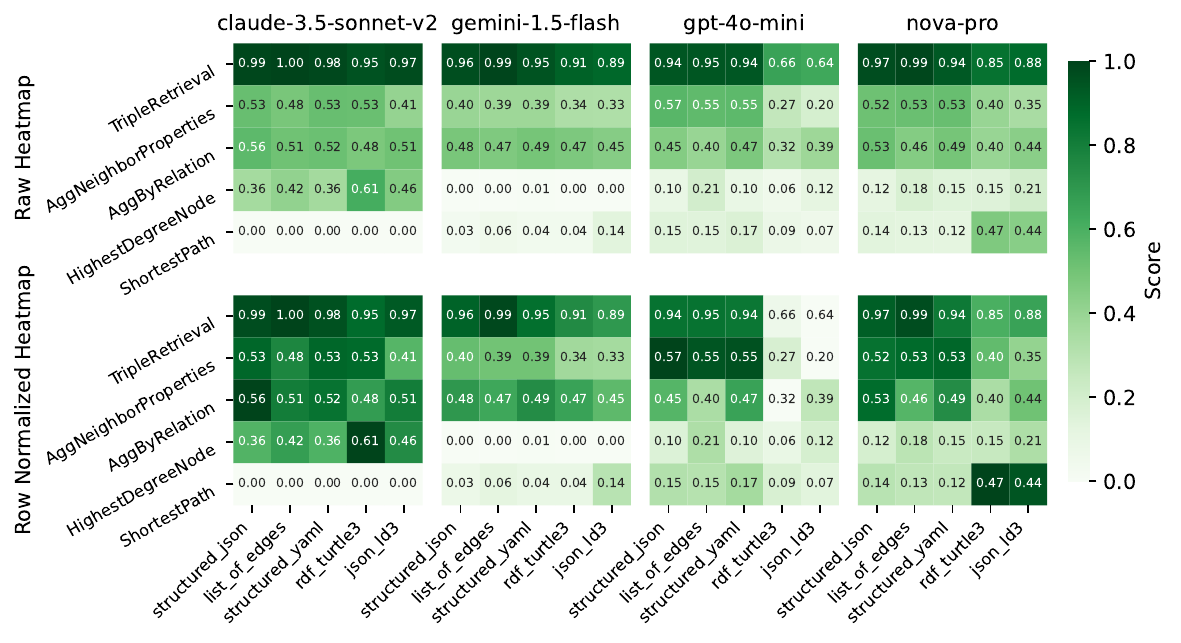}
    \caption{Heatmaps of the performance of various models. Each heatmap shows tasks as rows and textualize functions as columns. \textbf{(Top)} Heatmap colors as globally weighted from [0.0-1.0]. \textbf{(bottom)} heatmap colors normalized for each task [task minimum-task maximum]. The tasks are ordered from easiest overall to hardest. The textualization functions are ordered from best performing overall to worst. Additional models are in the appendix.}
    \label{fig:heatmaps}
\end{figure*}

\subsection{Models}

We evaluate seven different language models spanning different sizes and architectures. These models are \textbf{Llama 3.3-70B} \citep{Meta2024Llama33}, \textbf{Llama 3.2-1B} \citep{Meta2024Llama32Quantized}, \textbf{GPT-4o-Mini} \citep{OpenAI2024GPT4oMini}, \textbf{Claude-3.5-Sonnet} \citep{Anthropic2024Claude35Sonnet}, \textbf{Amazon Nova Lite} \citep{Intelligence2024}, \textbf{Amazon Nova Pro} \citep{Intelligence2024}, and \textbf{Gemini-1.5-Flash} \citep{geminiteam2024gemini15unlockingmultimodal}
This selection allows us to evaluate the effect of textualization strategies across a broad range of models.

\subsection{Evaluation Protocol}

To construct the benchmark and run the experiments, we do the following steps for each task.

\begin{enumerate}[noitemsep,nolistsep]
    \item Sample 100 subgraphs from $\mathcal{K}$ (Sec \ref{sec:sampling}) and pseudonymize each subgraph (Eq. \ref{eq:subgraph}).
    \item Generate questions and answers following task-specific protocols (Sec \ref{sec:queryconstruct}).
    \item Apply textualization strategies $f\in\mathcal{F}$ (Eq. \ref{eq:textualize})
    \item For each dataset, use the model to generate responses (Eq. \ref{eq:llm})
    \item Evaluate responses with exact match (Eq. \ref{eq:scoring})
\end{enumerate}

\section{Results}

The following section presents our results. Our main finding is the best overall textualization strategy is Structured JSON followed closely by List of Edges. However, there is a complex interplay between the textualization of the graph, the model and the task. This can be seen in the different performance patterns in Figure \ref{fig:heatmaps}. Therefore, developers must optimize textualization choice for their specific use case and model. We will present analysis on the high-level effect of textualization choice, how the selected models compare to each other, the effect of pseudonymization on suppressing memorized information, and the token efficiency of the different textualization strategies. 

The full results data is presented in Table \ref{tab:full_results_summary} in the appendix due to space constraints, and more tailored results are presented in this section. 

\subsection{Effect of Textualization Function}


While not all global results on textualization hold true for every model, there are some global patterns that we see. Figure \ref{fig:format_radar} gives a radar plot of performance by textualization function. Structured JSON performs best (0.42 average), followed by YAML and List-of-Edges, while RDF Turtle (0.35) and JSON-LD (0.34) perform worst. Part of the reason for this may be the more complex encoding strategies and use of URIs makes the format more difficult to parse. This may be further amplified by the fact that it dramatically increases the input token counts which may cause performance degradation on some of the models and tasks. 

Across all tasks, List-of-Edges and Structured-JSON perform quite well. List-of-Edges is commonly used, and thus may be the most common encoding format in instruction tuning data. On the aggregation tasks Structured YAML and Structured JSON outperform. This makes sense as these structures naturally aggregate related edges together. List-of-Edges only seems better on the global task of Highest Degree Node. This, too, makes sense, as the highest degree node will appear the most times in the list of edges format, but that is not guaranteed nor likely to be the case in the structured formats. 

\subsection{Model Performance}

\begin{figure}[t]
    \centering
    \begin{subfigure}{0.48\linewidth}
        \centering
        \includegraphics[width=\linewidth]{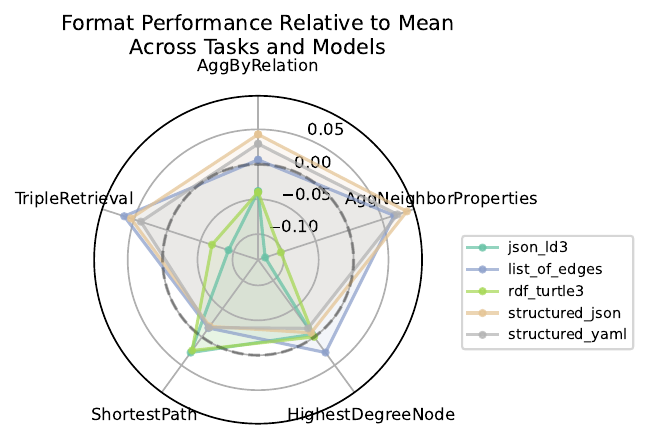}
        \caption{Textualization strategies comparison. }
        \label{fig:format_radar}
    \end{subfigure}
    \hfill
    \begin{subfigure}{0.48\linewidth}
        \centering
        \includegraphics[width=\linewidth]{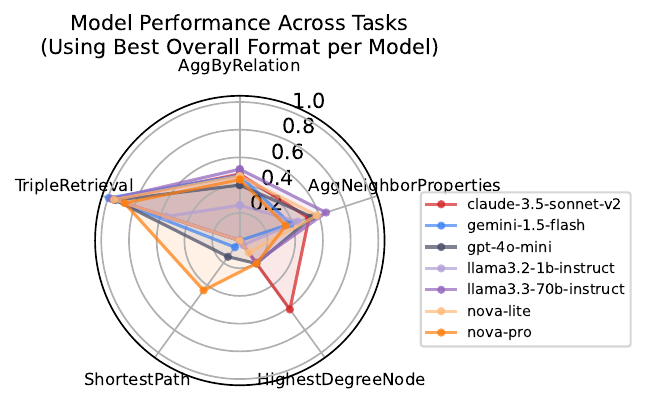}
        \caption{Model performance comparison.}
        \label{fig:model_radar}
    \end{subfigure}
    \caption{Performance analysis across tasks: (a) comparison of textualization strategies and (b) performance by model. The metric for (a) shows absolute difference in accuracy for each strategy compared to the mean for that task. The mean is shown as the dashed circle. }
    \label{fig:combined_radar}
\end{figure}

\begin{figure}
    \begin{minipage}{0.5\textwidth}
        \centering
        \includegraphics[width=\linewidth]{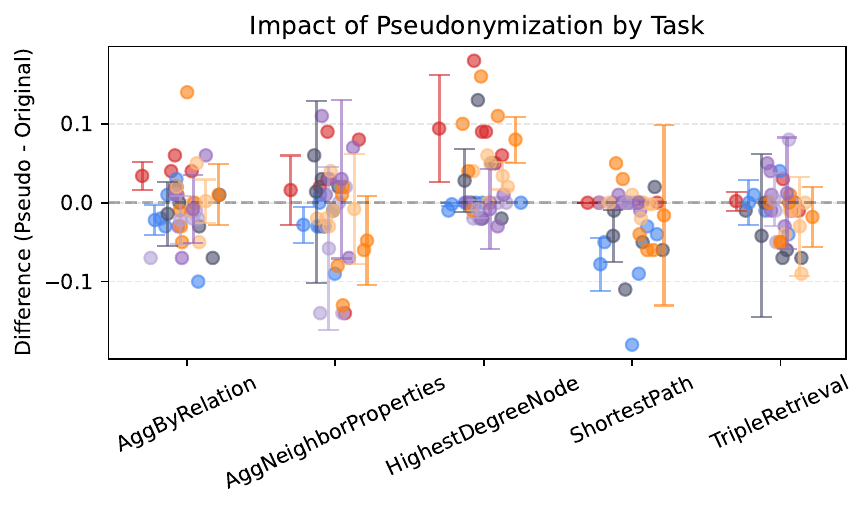}
        \caption{Impact of pseudonymization by task. Higher means that the model did better with pseudonymization. Each color represents a different model.}
        \label{fig:pseudo_by_task}
    \end{minipage}
    \hfill
    \begin{minipage}{0.46\textwidth}
        \input{tables/token_usage}
    \end{minipage}
\end{figure}

We can also use KG-LLM-Bench to compare the performance of various models. We plot the comparative performances of the seven models in Figure \ref{fig:model_radar}. To enable a fair comparison, we use data from the best performing overall textualization strategy for each model (Table \ref{tab:best_format_per_model}). We can see that the best performing approach is highly variable model to model. 

Overall, task performance and model rankings align with expectations. The easiest task is Triple Retrieval and the hardest task is Shortest Path. Surprisingly, Highest Degree task appeared to be significantly harder (most models less than 20\%) compared to the two aggregation tasks (most models scoring 40\%-60\%).

There were two notable outliers on performance. Nova-Pro scored by far the highest on the Shortest Path task, 47\% with RDF Turtle and 44\% with JSON-LD. The next best single Shortest Path result was gpt-4o-mini scoring 17\% with Structured YAML. Further analysis of the Shortest Path task can be found in Appendix \ref{app:shortest_path_analysis}. The other major outlier is Claude-3.5-Sonnet performance on the Highest Degree task. Sonnet received 61.5\% with RDF-Turtle and averaged 44.3\% over all formats. This is much better than the next best performer, Nova Pro at 16.2\%. Partially helped by these outlier abilities, Claude-3.5-Sonnet and Nova-Pro were the top overall models.

\subsection{Pseudonymization}

Pseudonymization shows minimal effect. This is likely because questions on sampled subgraphs of $\mathcal{K}$ already prevent reliance on memorized information since there is low chance of the memorized information being present (e.g. knowing the number of countries France borders is not helpful if $G$ only contains a subset of those edges). There is some limited evidence in Figure \ref{fig:pseudo_by_task} that pseudonymization actually helped on the Highest Degree Task. It may be that the model would erroneously guess based on memorized knowledge when it saw familiar entity names.

\subsection{Token Efficiency}

Textualization strategies can vary a lot in terms of token efficiency. List-of-Edges and and Structured YAML are the most token efficient with below 3000 tokens per prompt. JSON-LD was the least token efficient, taking over 13,000 tokens per prompt followed by RDF Turtle at ~8,000 tokens per prompt. Since RDF Turtle and JSON-LD are designed to be usable with semantic web technologies, they require complete and unambiguous specification of the schema. This results in many additional specifications like namespaces and URI encodings. We note that we even optimized some of these choices to reduce token usage. More naive encodings could use far more tokens.  

\subsection{Aggregation Performance Depends on Direction and Degree}

\begin{figure}[t]
    \centering
    \begin{minipage}{0.48\textwidth}
        \centering
        \includegraphics[width=\linewidth]{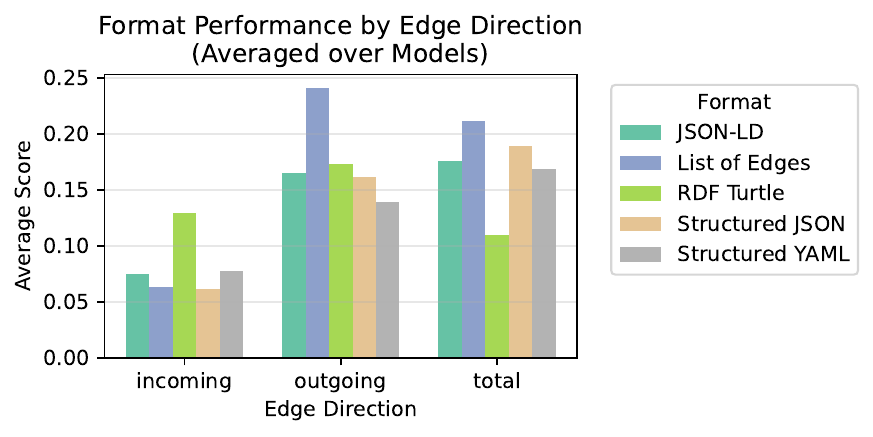}
        \captionof{figure}{Performance on highest degree task by aggregation direction. Aggregation over outgoing edges is easier in all formats due to locality of outgoing edges to each other.}
        \label{fig:highest_degree}
    \end{minipage}
    \hfill
    \begin{minipage}{0.48\textwidth}
        \centering
        \includegraphics[width=\linewidth]{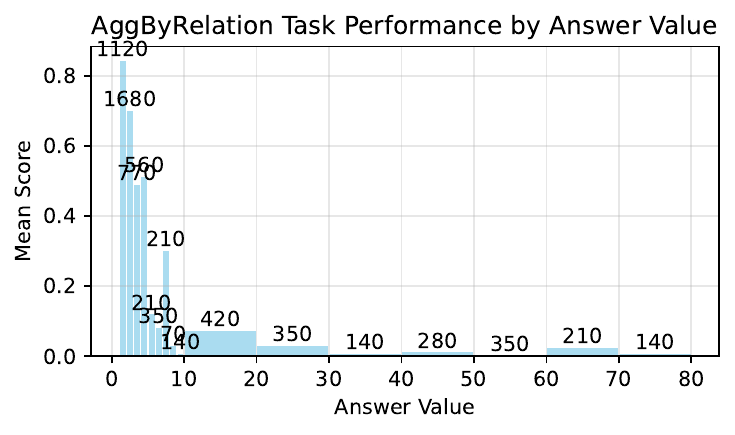}
        \captionof{figure}{Effect of aggregation size on accuracy in AggByRelation task. Numbers show the count of results in each bin. We note that these do not represent completely i.i.d. samples as the same base question will be presented many times.}
        \label{fig:agg_performance_by_degree}
    \end{minipage}
\end{figure}

We find that aggregation performance significantly differs by aggregation direction. Figure \ref{fig:highest_degree} shows the effect of aggregation direction when performing the HighestDegree task. The models do significantly better when predicting highest degree by outgoing edges than by incoming. In all the textualization formats (including our implementation of List of Edges), outgoing edges are listed next to each other. This makes it much easier for the model to aggregate over outgoing edges than over incoming ones. This difference is least pronounced in RDF Turtle. 

We also analyze how degree affects model performance on AggregationByRelation (Figure \ref{fig:agg_performance_by_degree}). For a single edge aggregation, the models answer correctly over 80\% of the time and remains above 50\% for aggregations up to degree 4. Beyond that, performance rapidly degrades to around 10\%. The steep drop indicates the models have significant room for improvement in aggregation capability.

\section{Future Work}

There are a many directions for future work based on KG-LLM-Bench. Our framework's modular design allows for easy extension to new tasks, graphs, models, and textualization strategies. 

In terms of research, we see two major directions to extend KG-LLM-Bench: Scale and Reasoning. In terms of scale, KG-LLM-Bench can be easily scaled by modifying the subgraph sampling parameters, and the source graph can be easily swapped should that become an issue. This enables the study of long-context reasoning over KGs. 

The other major direction would be to support studying of test-time reasoning on KGs. The generation of new queries is a bottleneck in developing reasoning models for LLMs. Since KG-LLM-Bench can continuously generate new queries, it could be a useful in training KG reasoning models.

\section{Conclusion}

In this work, we introduced KG-LLM-Bench, a comprehensive and extensible benchmark for evaluating how LLMs process and understand textualized knowledge graphs. Through extensive experiments across five distinct tasks, seven models, and five textualization strategies, we demonstrated that the choice of textualization strategy has a significant impact on model performance. While simpler formats like List-of-Edges and Structured JSON tend to perform well overall, the best performing format varies depending on model and task. 

 By understanding and improving how LLMs process structured knowledge, we can ultimately develop more reliable and effective knowledge-enhanced language models.

\section{Ethical Statement}

We see no immediate ethical issues with this work. The authors believe that more factual and trustworthy AI models are ethically desirable. However, this work can be used for enhancing AI capabilities, which could present other ethical ramifications. We encourage anyone using KG-LLM-Bench to consider the ethical impact of their work or applications.

\bibliography{kg_llm_bench}
\bibliographystyle{colm2025_conference}

\appendix

\section{Limitations}

Here we note some limitations of our experiments and framework. First, our evaluation uses only the WikiDataSets Countries knowledge graph. While this provides a controlled environment, the experiments should be expanded to other domains with different relationship types. Second, we use subgraphs of 200 edges to ensure reasonable context windows, which may not accurately reflect capabilities on smaller graphs, nor capture the challenges of reasoning over larger knowledge graphs. Finally, while we evaluate different textualization strategies, our experiments are conducted in English (though some entities are in other languages). 

There are a few current limitations in our framework. It currently only handles knowledge graphs with the defined triple structure. It cannot handle literal values (numbers or dates), temporal knowledge graphs, or any form of hypergraph (e.g. Wikidata edges can have qualifier edges). These would make good areas for future expansion. 

\section{Additonal Implementation Details}

\subsection{Evaluation Algorithm}

Algorithm \ref{alg:kg_llm_bench} gives the formal algorithm for KG-LLM-Bench constructin and evaluation. 

\begin{algorithm}
\DontPrintSemicolon
\SetAlgoVlined
\SetKwInOut{Input}{Input}
\SetKwInOut{Output}{Output}
\SetKwComment{tcp}{$\triangleright$}{}
\Input{
  $\mathcal{K}=(\mathcal{E}, \mathcal{R}, \mathcal{T})$: Knowledge graph \\
  $\mathcal{F}$: Textualization function set \\
  $\mathcal{T}$: Task set \\
  $\pi$: Language model \\
  $\hat{\mathcal{E}}$: Pseudonym entity set \\
  $N$: Number of instances
}
\Output{
  $\{s_{ijk}\}$: Evaluation scores where $i,j,k$ index tasks, textualization, and question instance
}

\For{$T \in \mathcal{T}$}{
   \tcc{Construct N instances}
   \For{$i \gets 1$ \KwTo $N$}{
       \tcc{Sample subgraph from knowledge graph}
       $G_i \sim \textit{subgraph}(\mathcal{K})$ \tcp*{Sec \ref{sec:sampling}}
       
       \tcc{Create anonymized version if requested}
       $\hat{G}_i \gets p(G_i, \hat{\mathcal{E}})$ \tcp*{Eq. \ref{eq:subgraph}}
       
       \tcc{Generate task-specific question and ground truth}
       $q_i \gets \mathcal{Q}_T(\hat{G}_i)$\;
       $a_i \gets \mathcal{A}_T(\hat{G}_i, q_i)$ \tcp*{Sec \ref{sec:queryconstruct}}
   }
   
   \tcc{Evaluate each textualization strategy}
   \For{$f \in \mathcal{F}$}{
       \For{$i \gets 1$ \KwTo $N$}{
           \tcc{Convert graph to textual format}
           $x_i \gets f(\hat{G}_i)$ \tcp*{Eq. \ref{eq:textualize}}
           
           \tcc{Query LLM with context and question}
           $\hat{y}_i \gets \pi(x_i, q_i)$ \tcp*{Eq. \ref{eq:llm}}
           
           \tcc{Evaluate response with exact match}
           $s_i \gets \mathcal{S}(\hat{y}_i, a_i)$ \tcp*{Eq. \ref{eq:scoring}}
       }
   }
}
\caption{KG-LLM-Bench}
\label{alg:kg_llm_bench}
\end{algorithm}

\subsection{Shortest Path Implementation}
\label{app:shortest_path}

We consider the shortest path between source entity $e_s$ and destination entity $e_d$ using edges in any direction. So for instance, we could use the edges $(e_s, r_1, e_1)$ and $(e_d, r_2, e_1)$ to form the path $[e_s, e_1, e_d]$. This makes the task potentially more difficult as it forces the model to rely on associations that appear in the reversed order from how they appear in the text representation $x_G$. 

To construct these questions, we take two of the seed entities used to construct the subgraph $G$ ($e_s$ and $e_d$) and set them as the source and destination nodes for the question. We then collect the set of all shortest paths $P = p_1, ..., p_k$ from $e_s$ to $e_d$ in $\mathcal{K}$. We use the set of all entities in $p_1$ as additional seed entities for $\textit{subgraph}(\mathcal{K})$ and further ensure that all edges in $p_1$ become part of $G$. This ensures that at least $p_1$ is present in the graph. Finally, we consider any answer in $P$ to be a valid answer.

\subsection{Aggregation by Relation Implementation}
\label{app:agg_by_relation}

Here we present additional details on the construction of AggByRelation questions. In our work, we use \texttt{COUNT} aggregation as we do not consider edges with literal expressions (e.g. numbers or dates), only other entities. 

We first compute the aggregation count for every possible question that could be constructed for $G$ (every direction, relation type, and anchor entity). Specifically
\begin{equation}
    Agg(s, r, dir) =
    \Bigg|\left\{ t = \begin{cases}
(s,r,e) & dir = 1 \\
(e,r,s) & dir = 0
\end{cases} 
\Big| e\in G_\mathcal{E}, t\in G_\mathcal{T} \right\}\Bigg|    
\end{equation}

We then collect the set of possible answers $A=\left\{Agg(s,r,dir)|s\in G_\mathcal{E}, r\in G_\mathcal{R}, dir\in[1,0]\right\}$ and randomly select an answer $a\sim A$. After selecting the desired answer, we finally sample the $s$, $r$, and $dir$ that would give answer $a$: $(s,r,dir)\sim\left\{(s,r,dir)| Agg(s,r,dir)=a\right\}$.

\subsection{Aggregation of Neighbor Properties Implementation}
\label{app:agg_neighbor_property}
To construct these questions we follow the same procedure as the previous task but use a different aggregation formula. 

\begin{equation}
    Agg(s, r) =
    \Bigg|\Big\{e_1 \in G_\mathcal{E} \Big| \exists t_1,t_2 \in G_\mathcal{T}:
    t_1\in\{(s,\_,e_1),(e_1,\_,s)\} \land t_2=(e_1,r,\_)\Big\}\Bigg|
\end{equation}

``$\_$'' indicates wildcards that could match any relation or entity that makes the edge valid in $G_\mathcal{T}$

\subsection{Sampling Parameters}
\label{appendix:sampling}

Table \ref{tab:sampling_params} details the sampling parameters used for each task in our benchmark. 

\begin{table*}[h]
\centering
\begin{tabular}{lrrrrr}
\hline
\textbf{Task} & \textbf{Instances} & \textbf{Seed Entities} & \textbf{Max Edges} & \textbf{Sample Radius} & \textbf{Min Degree} \\
\hline
Triple Retrieval & 100 & 10 & 200 & 1 & 1 \\
Shortest Path & 100 & 10 & 200 & 1 & 1 \\
Highest Degree Node & 100 & 10 & 200 & 1 & 1 \\
Agg by Relation & 100 & 1 & 200 & 2 & 2 \\
Agg of Neighbor Properties & 100 & 1 & 200 & 2 & 2 \\
\hline
\end{tabular}
\caption{Sampling parameters used for each task. Min degree filter is applied before the max edge constraint.}
\label{tab:sampling_params}
\end{table*}

\subsection{Pseudonymization}

For pseudonymization, we use a pre-defined set of country pseudonyms stored in csv file. These pseudonyms are designed to maintain the semantic naturalness of the graph while preventing the model from leveraging pre-trained knowledge about real countries. 

We use a fake names generator to generate the first 100 fake country names and then use claude-3.5-sonnet to generate an additional 600 of a similar style. 

\section{Shortest Path Task Additional Analysis}
\label{app:shortest_path_analysis}

We provide here additional analysis for the shortest path task. 

\subsection{Flexible Matching on Shortest Path}

\begin{figure*}[t]
    \centering
    \includegraphics[width=0.98\linewidth]{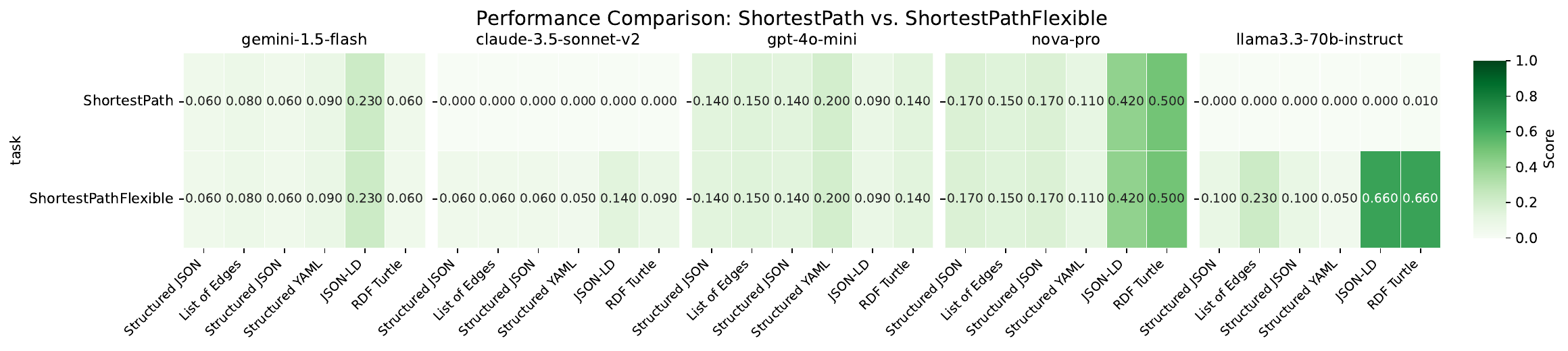}
    \caption{Heatmap of ShortestPath performance when allowing more flexible matching vs exact match. Models that ignored formatting instructions to output more chain-of-thought tokens get higher performance under this scoring approach.}
    \label{fig:flexible_path}
\end{figure*}

A few of the models had a tendency to ignore formatting instructions and used extensive chain-of-thought. As a result, we present an analysis of their output under "flexible" scoring criteria. This is presented in Figure \ref{fig:flexible_path}. The models that followed formatting guidelines strictly have no difference in their scores. Models like Llama3.3-70b and Claude-3.5-Sonnet see improvement with the flexible scoring. Interestingly, Llama3.3-70b also has very high performance on the JSON-LD and RDF Turtle formats, just like Nova Pro. Under flexible scoring, Claude-3.5-Sonnet also does slightly better with theses formats. This shows that there may be a more fundamental reason why JSON-LD and RDF Turtle are better for this task. 

\subsection{Path Length}

\begin{figure*}[h]
    \centering
    \includegraphics[width=0.7\linewidth]{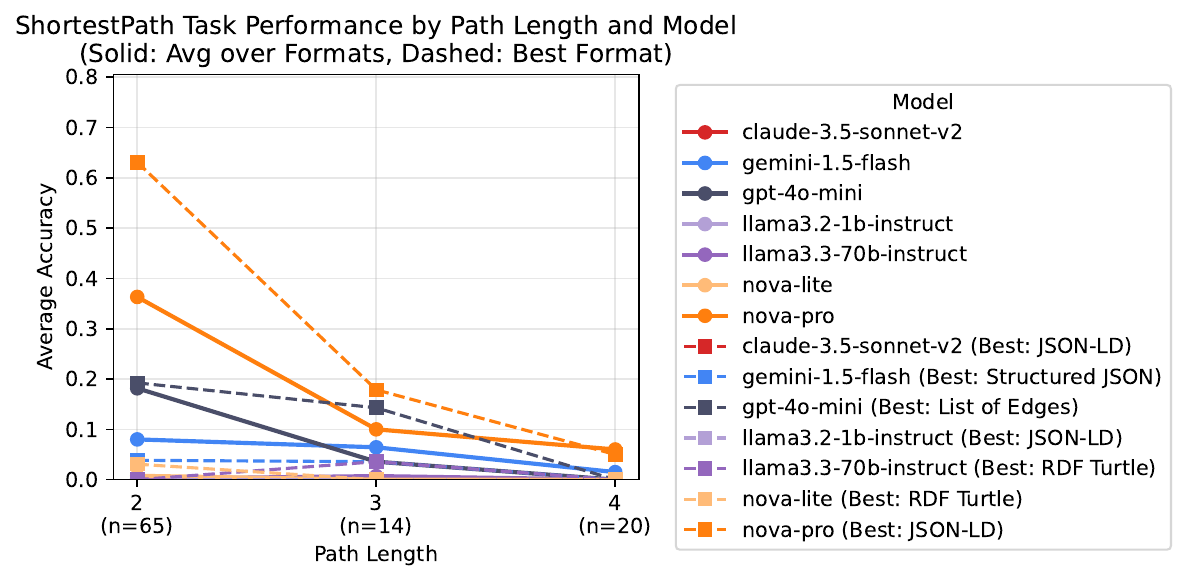}
    \caption{Accuracy by path length. Solid lines are mean performance. Dashed lines are for the best performing textualize format. Path length 5 has only a single question so was excluded from this analysis.}
    \label{fig:acc_by_path_length}
\end{figure*}

In terms of the effect of path length, we see a clear decrease in performance when path length increases. Figure \ref{fig:acc_by_path_length} shows the performance for each model in terms of the true shortest path length. There is a clear decreasing trend in performance where the shorter the shortest path is, the easier it is for models to find it. 

\begin{figure*}
    \centering
    \includegraphics[width=0.7\linewidth]{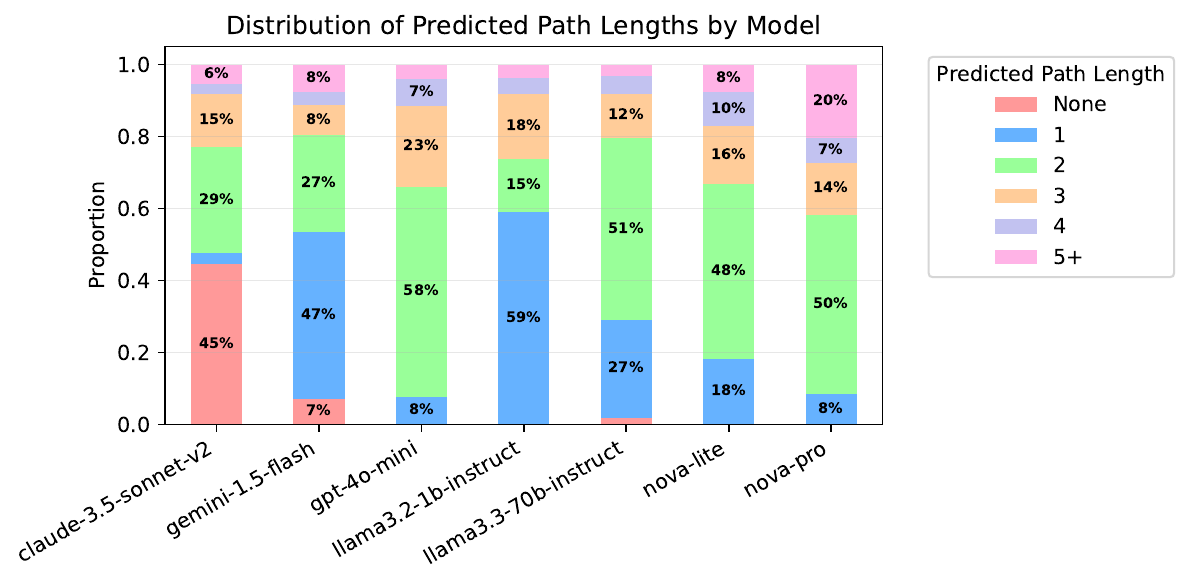}
    \caption{Distribution of predicted path lengths for each model. "None" indicates an empty path or refusal to generate a valid output path.}
    \label{fig:path_length_distribution}
\end{figure*}

\begin{figure*}
    \centering
    \includegraphics[width=0.5\linewidth]{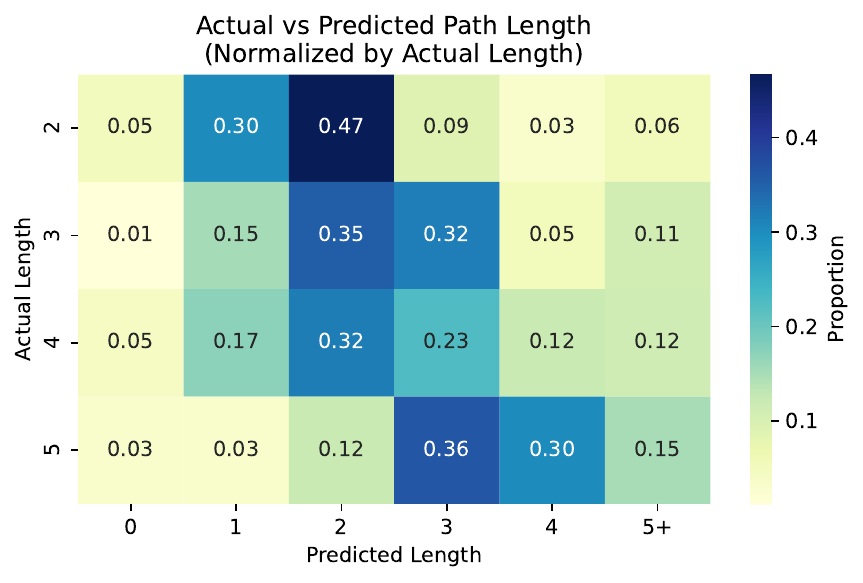}
    \caption{Heatmap of length of predicted paths vs the actual shortest path length. Cases where no path is predicted are excluded.}
    \label{fig:path_length_vs_predicted_length}
\end{figure*}

Figure \ref{fig:path_length_distribution} gives the distribution of predicted path lengths for each model. There are clear variations in model prediction patterns. For a large proportion of questions, Claude-3.5-Sonnet actually predicts that no path exists at all. Another common error is predicting an immediate path between the source and destination. This can often be caused by shared properties, but then the model skips the intermediary entity. Figure \ref{fig:path_length_vs_predicted_length} shows a heatmap of predicted lengths vs actual lengths (excluding cases where no path is predicted). While predicted lengths trend upwards with increasing actual shortest paths, there is also a clear trend of under predicting. Most predicted lengths are less than the actual length. This means that the models are often outputting paths with hallucinated edges. 

\newpage
\section{Examples of Text Formats}
\label{app:text_formats}

\input{tables/example_formats}

\section{Full Results}

We present the full results and data over the following pages. 

\subsection{Heatmap Results}

\begin{figure*}[h]
    \centering
    \includegraphics[width=\linewidth]{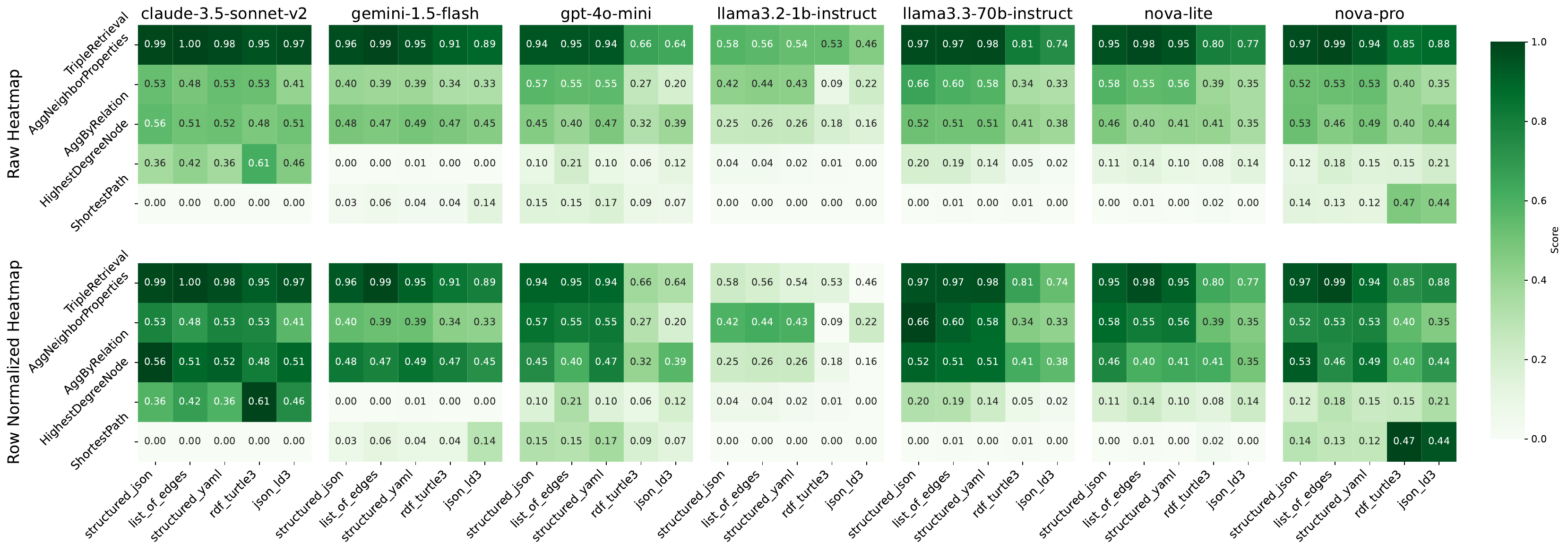}
    \caption{Heatmaps of the performance of various models. Each heatmap shows tasks as rows and textualize functions as columns. The top row of the grid shows the heatmap colors as globally weighted from [0.0-1.0]. The bottom grid shows heatmap colors normalized for each task [task minimum-task maximum].}
    \label{fig:heatmaps_all}
\end{figure*}

Figure \ref{fig:heatmaps_all} presents the heatmap data for all models.

\onecolumn
\input{tables/results}
\twocolumn

\end{document}

%% file: tables/combined_dataset_table_and_best_format_table.tex
\begin{table}[t]
\begin{minipage}{0.5\textwidth}
\centering
\begin{tabular}{lr}
\hline
\textbf{Statistic} & \textbf{Count} \\
\hline
Core Entities & 3,552 \\
Attribute Entities & 27,226 \\
Core Relations & 49 \\
Attribute Relations & 162 \\
Core Facts & 11,361 \\
Attribute Facts & 51,952 \\
\hline
\end{tabular}
\caption{WikiDataSets Countries. Core entities are countries. Attribute entities are related concepts such as languages or significant events. Core relations/facts involve only country-to-country relationships, while attribute relations/facts connect countries to their attributes.}
\label{tab:dataset}
\end{minipage}
\hfill
\begin{minipage}{0.46\textwidth}
\centering
\input{tables/best_format_for_model}
\caption{Best Textualization Strategy for Each Model}
\label{tab:best_format_per_model}
\end{minipage}
\end{table}

%% file: tables/best_format_for_model.tex
\begin{tabular}{ll}
\toprule
Model & Best $f$ \\
\midrule
claude-3.5-sonnet-v2 & RDF Turtle \\
gemini-1.5-flash & List of Edges \\
gpt-4o-mini & List of Edges \\
llama3.2-1b-instruct & List of Edges \\
llama3.3-70b-instruct & Structured JSON \\
nova-lite & Structured JSON \\
nova-pro & JSON-LD \\
\bottomrule
\end{tabular}

%% file: tables/token_usage.tex
\captionof{table}{Average Input Token Usage by Format}
\begin{tabular}{lr}
\toprule
Textualizer & Mean Input Tokens \\
\midrule
List of Edges & 2644.8 $\pm$ 390.4 \\
Structured JSON & 4504.7 $\pm$ 1123.2 \\
Structured YAML & 2903.1 $\pm$ 655.9 \\
RDF Turtle & 8171.1 $\pm$ 2284.6 \\
JSON-LD & 13503.4 $\pm$ 3611.2 \\
\midrule
Overall & 6345.4 $\pm$ 1613.1 \\
\bottomrule
\end{tabular}

%% file: tables/example_formats.tex
\subsection{List of Edges}

\begin{lstlisting}
Your job is to answer questions using the following knowledge graph. The knowledge graph is presented as a structured JSON format. Each entity is a key, and the value is a dictionary of relations and objects.. You must rely exclusively on the information presented in the Knowledge Graph to answer questions. If the answer includes entities, always respond using the entity label rather than entity ID (if applicable).

Knowledge Graph:
Edges: [
(Andhra Pradesh, language used, Telugu),
(Andhra Pradesh, language used, Marathi),
(Andhra Pradesh, language used, Odia),
(Guatemala, capital of, Federal Republic of Central America),
(Guatemala, diplomatic relation, European Union),
(Brunei, member of, World Trade Organization),
(Brunei, member of, International Hydrographic Organization),
(South Korea, diplomatic relation, Ukraine),
(South Korea, diplomatic relation, Colombia),
(South Korea, member of, G20)
]
\end{lstlisting}

\subsection{Structured JSON}

\begin{lstlisting}
Your job is to answer questions using the following knowledge graph. The knowledge graph is presented as a list of directed edges of the form (subject, relation, object). You must rely exclusively on the information presented in the Knowledge Graph to answer questions. If the answer includes entities, always respond using the entity label rather than entity ID (if applicable).
Knowledge Graph:
{
    "Andhra Pradesh": {
        "language used": [
            "Telugu",
            "Marathi",
            "Odia"
        ]
    },
    "Guatemala": {
        "capital of": [
            "Federal Republic of Central America"
        ],
        "diplomatic relation": [
            "European Union"
        ]
    },
    "Brunei": {
        "member of": [
            "World Trade Organization",
            "International Hydrographic Organization"
        ]
    },
    "South Korea": {
        "diplomatic relation": [
            "Ukraine",
            "Colombia"
        ],
        "member of": [
            "G20"
        ]
    }
}
\end{lstlisting}

\subsection{Structured YAML}

\begin{lstlisting}
Your job is to answer questions using the following knowledge graph. The knowledge graph is presented as a structured YAML format. Each entity is a key, and the value is a dictionary of relations and objects.. You must rely exclusively on the information presented in the Knowledge Graph to answer questions. If the answer includes entities, always respond using the entity label rather than entity ID (if applicable).

Knowledge Graph:
Andhra Pradesh:
  language used:
    - Telugu
    - Marathi
    - Odia

Guatemala:
  capital of:
    - Federal Republic of Central America
  diplomatic relation:
    - European Union

Brunei:
  member of:
    - World Trade Organization
    - International Hydrographic Organization

South Korea:
  diplomatic relation:
    - Ukraine
    - Colombia
  member of:
    - G20
\end{lstlisting}

\subsection{RDF Turtle}

\begin{lstlisting}
Your job is to answer questions using the following knowledge graph. The knowledge graph is presented as RDF Turtle format using node IDs and relation IDs.. You must rely exclusively on the information presented in the Knowledge Graph to answer questions. If the answer includes entities, always respond using the entity label rather than entity ID (if applicable).

Knowledge Graph:
@prefix ex: <http://example.org/countries#> .
@prefix rdf: <http://www.w3.org/1999/02/22-rdf-syntax-ns#> .
@prefix rdfs: <http://www.w3.org/2000/01/rdf-schema#> .

ex:R1 a rdf:Property ;
    rdfs:label "language used" .

ex:R2 a rdf:Property ;
    rdfs:label "member of" .

ex:R3 a rdf:Property ;
    rdfs:label "diplomatic relation" .

ex:R4 a rdf:Property ;
    rdfs:label "capital of" .

ex:1 a ex:Country ;
    rdfs:label "Andhra Pradesh" ;
    ex:R1 ex:101, ex:102, ex:103 .

ex:101 a ex:Language ;
    rdfs:label "Telugu" .

ex:102 a ex:Language ;
    rdfs:label "Marathi" .

ex:103 a ex:Language ;
    rdfs:label "Odia" .

ex:2 a ex:Country ;
    rdfs:label "Guatemala" ;
    ex:R4 ex:201 ;
    ex:R3 ex:202 .

ex:201 a ex:Country ;
    rdfs:label "Federal Republic of Central America" .

ex:202 a ex:Organization ;
    rdfs:label "European Union" .

ex:3 a ex:Country ;
    rdfs:label "Brunei" ;
    ex:R2 ex:301, ex:302 .

ex:301 a ex:Organization ;
    rdfs:label "World Trade Organization" .

ex:302 a ex:Organization ;
    rdfs:label "International Hydrographic Organization" .

ex:4 a ex:Country ;
    rdfs:label "South Korea" ;
    ex:R3 ex:401, ex:402 ;
    ex:R2 ex:403 .

ex:401 a ex:Country ;
    rdfs:label "Ukraine" .

ex:402 a ex:Country ;
    rdfs:label "Colombia" .

ex:403 a ex:Organization ;
    rdfs:label "G20" .
\end{lstlisting}

\subsection{JSON-LD}

\begin{lstlisting}
"Your job is to answer questions using the following knowledge graph. The knowledge graph is presented as JSON-LD format using node IDs and relation IDs.. You must rely exclusively on the information presented in the Knowledge Graph to answer questions. If the answer includes entities, always respond using the entity label rather than entity ID (if applicable).

Knowledge Graph:
{
  "@context": {
    "@context": {
      "ex": "http://example.org/countries#",
      "label": "rdfs:label",
      "rdf": "http://www.w3.org/1999/02/22-rdf-syntax-ns#",
      "rdfs": "http://www.w3.org/2000/01/rdf-schema#",
      "type": "@type"
    }
  },
  "@graph": [
    {
      "@id": "ex:R1",
      "label": "language used",
      "type": "rdf:Property"
    },
    {
      "@id": "ex:R3",
      "label": "diplomatic relation",
      "type": "rdf:Property"
    },
    {
      "@id": "ex:R4",
      "label": "capital of",
      "type": "rdf:Property"
    },
    {
      "@id": "ex:1",
      "type": "ex:Country",
      "label": "Andhra Pradesh",
      "ex:R1": [
        { "@id": "ex:101" },
        { "@id": "ex:102" },
        { "@id": "ex:103" }
      ]
    },
    {
      "@id": "ex:101",
      "type": "ex:Language",
      "label": "Telugu"
    },
    {
      "@id": "ex:102",
      "type": "ex:Language",
      "label": "Marathi"
    },
    {
      "@id": "ex:103",
      "type": "ex:Language",
      "label": "Odia"
    },
    {
      "@id": "ex:2",
      "type": "ex:Country",
      "label": "Guatemala",
      "ex:R4": { "@id": "ex:201" },
      "ex:R3": { "@id": "ex:202" }
    },
    {
      "@id": "ex:201",
      "type": "ex:Country",
      "label": "Federal Republic of Central America"
    },
    {
      "@id": "ex:202",
      "type": "ex:Organization",
      "label": "European Union"
    }
  ]
}
\end{lstlisting}

%% file: tables/results.tex

\begin{longtable}{p{1.5cm}lcccccc}
\caption{Full Results Summary by Format and Model} \\
\label{tab:full_results_summary} \\
\toprule
Format & Model & \makecell{Agg by\\ Relation} & \makecell{Agg Neighbor\\ Properties} & \makecell{Highest\\ Degree} & \makecell{Shortest\\ Path} & \makecell{Triple\\ Retrieval} & Overall \\
\midrule
\endfirsthead
\multicolumn{8}{c}{\tablename\ \thetable\ -- Continued from previous page} \\
\toprule
Format & Model & \makecell{Agg by\\ Relation} & \makecell{Agg Neighbor\\ Properties} & \makecell{Highest\\ Degree} & \makecell{Shortest\\ Path} & \makecell{Triple\\ Retrieval} & Overall \\
\midrule
\endhead
\midrule
\multicolumn{8}{r}{Continued on next page} \\
\endfoot
\bottomrule
\endlastfoot
\multirow{2}{=}{\rotatebox[origin=c]{90}{List of Edges}} & \textbf{claude-3.5-sonnet-v2} & 0.490 & 0.440 & 0.370 & 0.000 & 1.000 & 0.460 \\
 & \quad +pseudo & 0.530 & 0.530 & 0.460 & 0.000 & 1.000 & 0.504 \\
 & \textbf{gemini-1.5-flash} & 0.520 & 0.430 & 0.000 & 0.080 & 1.000 & 0.406 \\
 & \quad +pseudo & 0.420 & 0.340 & 0.000 & 0.040 & 0.990 & 0.358 \\
 & \textbf{gpt-4o-mini} & 0.400 & 0.520 & 0.140 & 0.150 & 0.980 & 0.438 \\
 & \quad +pseudo & 0.400 & 0.580 & 0.270 & 0.140 & 0.910 & 0.460 \\
 & \textbf{llama3.2-1b-instruct} & 0.250 & 0.430 & 0.050 & 0.000 & 0.560 & 0.258 \\
 & \quad +pseudo & 0.260 & 0.450 & 0.030 & 0.000 & 0.560 & 0.260 \\
 & \textbf{llama3.3-70b-instruct} & 0.540 & 0.590 & 0.200 & 0.000 & 0.970 & 0.460 \\
 & \quad +pseudo & 0.470 & 0.620 & 0.180 & 0.010 & 0.980 & 0.452 \\
 & \textbf{nova-lite} & 0.390 & 0.560 & 0.120 & 0.000 & 0.990 & 0.412 \\
 & \quad +pseudo & 0.410 & 0.540 & 0.160 & 0.010 & 0.980 & 0.420 \\
 & \textbf{nova-pro} & 0.460 & 0.520 & 0.130 & 0.150 & 0.990 & 0.450 \\
 & \quad +pseudo & 0.450 & 0.530 & 0.230 & 0.110 & 0.990 & 0.462 \\
\midrule
 & \textbf{Format Overall} & 0.436 & 0.499 & 0.144 & 0.054 & 0.927 & 0.412 \\
 & \quad +pseudo & 0.420 & 0.513 & 0.190 & 0.044 & 0.916 & 0.417 \\
\midrule
\multirow{2}{=}{\rotatebox[origin=c]{90}{Structured JSON}} & \textbf{claude-3.5-sonnet-v2} & 0.550 & 0.520 & 0.330 & 0.000 & 0.990 & 0.478 \\
 & \quad +pseudo & 0.560 & 0.540 & 0.390 & 0.000 & 0.990 & 0.496 \\
 & \textbf{gemini-1.5-flash} & 0.500 & 0.410 & 0.000 & 0.060 & 0.960 & 0.386 \\
 & \quad +pseudo & 0.470 & 0.380 & 0.000 & 0.010 & 0.970 & 0.366 \\
 & \textbf{gpt-4o-mini} & 0.490 & 0.560 & 0.100 & 0.140 & 0.950 & 0.448 \\
 & \quad +pseudo & 0.420 & 0.580 & 0.100 & 0.160 & 0.940 & 0.440 \\
 & \textbf{llama3.2-1b-instruct} & 0.260 & 0.440 & 0.040 & 0.000 & 0.540 & 0.256 \\
 & \quad +pseudo & 0.240 & 0.410 & 0.040 & 0.000 & 0.620 & 0.262 \\
 & \textbf{llama3.3-70b-instruct} & 0.530 & 0.600 & 0.190 & 0.000 & 0.980 & 0.460 \\
 & \quad +pseudo & 0.500 & 0.710 & 0.200 & 0.000 & 0.970 & 0.476 \\
 & \textbf{nova-lite} & 0.490 & 0.590 & 0.100 & 0.000 & 0.960 & 0.428 \\
 & \quad +pseudo & 0.440 & 0.580 & 0.120 & 0.000 & 0.950 & 0.418 \\
 & \textbf{nova-pro} & 0.550 & 0.580 & 0.100 & 0.170 & 0.970 & 0.474 \\
 & \quad +pseudo & 0.500 & 0.450 & 0.140 & 0.110 & 0.970 & 0.434 \\
\midrule
 & \textbf{Format Overall} & 0.481 & 0.529 & 0.123 & 0.053 & 0.907 & 0.419 \\
 & \quad +pseudo & 0.447 & 0.521 & 0.141 & 0.040 & 0.916 & 0.413 \\
\midrule
\multirow{2}{=}{\rotatebox[origin=c]{90}{Structured YAML}} & \textbf{claude-3.5-sonnet-v2} & 0.500 & 0.600 & 0.320 & 0.000 & 0.990 & 0.482 \\
 & \quad +pseudo & 0.540 & 0.460 & 0.410 & 0.000 & 0.980 & 0.478 \\
 & \textbf{gemini-1.5-flash} & 0.490 & 0.400 & 0.010 & 0.090 & 0.950 & 0.388 \\
 & \quad +pseudo & 0.500 & 0.370 & 0.000 & 0.000 & 0.950 & 0.364 \\
 & \textbf{gpt-4o-mini} & 0.490 & 0.540 & 0.070 & 0.200 & 0.940 & 0.448 \\
 & \quad +pseudo & 0.460 & 0.570 & 0.120 & 0.140 & 0.940 & 0.446 \\
 & \textbf{llama3.2-1b-instruct} & 0.290 & 0.430 & 0.030 & 0.000 & 0.560 & 0.262 \\
 & \quad +pseudo & 0.220 & 0.430 & 0.010 & 0.000 & 0.510 & 0.234 \\
 & \textbf{llama3.3-70b-instruct} & 0.510 & 0.550 & 0.150 & 0.000 & 0.960 & 0.434 \\
 & \quad +pseudo & 0.500 & 0.620 & 0.120 & 0.000 & 1.000 & 0.448 \\
 & \textbf{nova-lite} & 0.410 & 0.540 & 0.100 & 0.000 & 0.950 & 0.400 \\
 & \quad +pseudo & 0.410 & 0.580 & 0.100 & 0.000 & 0.950 & 0.408 \\
 & \textbf{nova-pro} & 0.510 & 0.520 & 0.070 & 0.110 & 0.930 & 0.428 \\
 & \quad +pseudo & 0.480 & 0.540 & 0.230 & 0.140 & 0.940 & 0.466 \\
\midrule
 & \textbf{Format Overall} & 0.457 & 0.511 & 0.107 & 0.057 & 0.897 & 0.406 \\
 & \quad +pseudo & 0.444 & 0.510 & 0.141 & 0.040 & 0.896 & 0.406 \\
\midrule
\multirow{2}{=}{\rotatebox[origin=c]{90}{RDF Turtle}} & \textbf{claude-3.5-sonnet-v2} & 0.450 & 0.510 & 0.590 & 0.000 & 0.940 & 0.498 \\
 & \quad +pseudo & 0.510 & 0.540 & 0.640 & 0.000 & 0.970 & 0.532 \\
 & \textbf{gemini-1.5-flash} & 0.460 & 0.330 & 0.000 & 0.060 & 0.890 & 0.348 \\
 & \quad +pseudo & 0.490 & 0.340 & 0.000 & 0.030 & 0.930 & 0.358 \\
 & \textbf{gpt-4o-mini} & 0.310 & 0.270 & 0.070 & 0.140 & 0.690 & 0.296 \\
 & \quad +pseudo & 0.330 & 0.260 & 0.050 & 0.030 & 0.630 & 0.260 \\
 & \textbf{llama3.2-1b-instruct} & 0.190 & 0.160 & 0.010 & 0.000 & 0.530 & 0.178 \\
 & \quad +pseudo & 0.180 & 0.020 & 0.010 & 0.000 & 0.530 & 0.148 \\
 & \textbf{llama3.3-70b-instruct} & 0.400 & 0.380 & 0.050 & 0.010 & 0.790 & 0.326 \\
 & \quad +pseudo & 0.410 & 0.310 & 0.050 & 0.000 & 0.840 & 0.322 \\
 & \textbf{nova-lite} & 0.410 & 0.400 & 0.060 & 0.030 & 0.820 & 0.344 \\
 & \quad +pseudo & 0.400 & 0.370 & 0.110 & 0.010 & 0.780 & 0.334 \\
 & \textbf{nova-pro} & 0.330 & 0.440 & 0.100 & 0.500 & 0.880 & 0.450 \\
 & \quad +pseudo & 0.470 & 0.360 & 0.210 & 0.440 & 0.830 & 0.462 \\
\midrule
 & \textbf{Format Overall} & 0.364 & 0.356 & 0.126 & 0.106 & 0.791 & 0.349 \\
 & \quad +pseudo & 0.399 & 0.314 & 0.153 & 0.073 & 0.787 & 0.345 \\
\midrule
\multirow{2}{=}{\rotatebox[origin=c]{90}{JSON-LD}} & \textbf{claude-3.5-sonnet-v2} & 0.500 & 0.370 & 0.370 & 0.000 & 0.980 & 0.444 \\
 & \quad +pseudo & 0.520 & 0.450 & 0.550 & 0.000 & 0.970 & 0.498 \\
 & \textbf{gemini-1.5-flash} & 0.460 & 0.330 & 0.000 & 0.230 & 0.910 & 0.386 \\
 & \quad +pseudo & 0.440 & 0.330 & 0.000 & 0.050 & 0.870 & 0.338 \\
 & \textbf{gpt-4o-mini} & 0.380 & 0.210 & 0.130 & 0.090 & 0.670 & 0.296 \\
 & \quad +pseudo & 0.390 & 0.180 & 0.110 & 0.040 & 0.600 & 0.264 \\
 & \textbf{llama3.2-1b-instruct} & 0.170 & 0.290 & 0.000 & 0.000 & 0.460 & 0.184 \\
 & \quad +pseudo & 0.150 & 0.150 & 0.000 & 0.000 & 0.450 & 0.150 \\
 & \textbf{llama3.3-70b-instruct} & 0.350 & 0.320 & 0.020 & 0.000 & 0.760 & 0.290 \\
 & \quad +pseudo & 0.410 & 0.330 & 0.020 & 0.000 & 0.730 & 0.298 \\
 & \textbf{nova-lite} & 0.330 & 0.360 & 0.110 & 0.000 & 0.820 & 0.324 \\
 & \quad +pseudo & 0.380 & 0.340 & 0.170 & 0.000 & 0.730 & 0.324 \\
 & \textbf{nova-pro} & 0.440 & 0.380 & 0.210 & 0.420 & 0.900 & 0.470 \\
 & \quad +pseudo & 0.440 & 0.320 & 0.200 & 0.470 & 0.850 & 0.456 \\
\midrule
 & \textbf{Format Overall} & 0.376 & 0.323 & 0.120 & 0.106 & 0.786 & 0.342 \\
 & \quad +pseudo & 0.390 & 0.300 & 0.150 & 0.080 & 0.743 & 0.333 \\
\midrule
\multirow{2}{=}{\rotatebox[origin=c]{90}{All Formats}} & \textbf{claude-3.5-sonnet-v2} & 0.498 & 0.488 & 0.396 & 0.000 & 0.980 & 0.472 \\
 & \quad +pseudo & 0.532 & 0.504 & 0.490 & 0.000 & 0.982 & 0.502 \\
 & \textbf{gemini-1.5-flash} & 0.486 & 0.380 & 0.002 & 0.104 & 0.942 & 0.383 \\
 & \quad +pseudo & 0.464 & 0.352 & 0.000 & 0.026 & 0.942 & 0.357 \\
 & \textbf{gpt-4o-mini} & 0.414 & 0.420 & 0.102 & 0.144 & 0.846 & 0.385 \\
 & \quad +pseudo & 0.400 & 0.434 & 0.130 & 0.102 & 0.804 & 0.374 \\
 & \textbf{llama3.2-1b-instruct} & 0.232 & 0.350 & 0.026 & 0.000 & 0.530 & 0.228 \\
 & \quad +pseudo & 0.210 & 0.292 & 0.018 & 0.000 & 0.534 & 0.211 \\
 & \textbf{llama3.3-70b-instruct} & 0.466 & 0.488 & 0.122 & 0.002 & 0.892 & 0.394 \\
 & \quad +pseudo & 0.458 & 0.518 & 0.114 & 0.002 & 0.904 & 0.399 \\
 & \textbf{nova-lite} & 0.406 & 0.490 & 0.098 & 0.006 & 0.908 & 0.382 \\
 & \quad +pseudo & 0.408 & 0.482 & 0.132 & 0.004 & 0.878 & 0.381 \\
 & \textbf{nova-pro} & 0.458 & 0.488 & 0.122 & 0.270 & 0.934 & 0.454 \\
 & \quad +pseudo & 0.468 & 0.440 & 0.202 & 0.254 & 0.916 & 0.456 \\
 \midrule
 & \textbf{Overall Score} & 0.423 & 0.443 & 0.124 & 0.075 & 0.862 & 0.385 \\
 & \quad +pseudo & 0.420 & 0.432 & 0.155 & 0.055 & 0.851 & 0.383 \\
\bottomrule
\end{longtable}